\documentclass{article}

\usepackage{arxiv} 
\usepackage{amsmath} 
\usepackage{graphicx}
\usepackage{subfig,caption}
\usepackage[utf8]{inputenc} % allow utf-8 input
\usepackage[T1]{fontenc}    % use 8-bit T1 fonts
\usepackage{hyperref}       % hyperlinks
\usepackage{url}            % simple URL typesetting
\usepackage{booktabs}       % professional-quality tables
\usepackage{amsfonts}       % blackboard math symbols
\usepackage{nicefrac}       % compact symbols for 1/2, etc.
\usepackage{microtype}      % microtypography
\usepackage{lipsum}
\usepackage[ruled, linesnumbered]{algorithm2e}
\usepackage{gensymb}

\title{Human-Expert-Level Brain Tumor Detection Using Deep Learning with Data Distillation and Augmentation}

\author{
	Diyuan~Lu\thanks{This work has been submitted to the IEEE for possible publication. Copyright may be transferred without notice, after which this version may no longer be accessible.
	}\\
	Frankfurt Institute for \\
	Advanced Studies (FIAS)\\
	Frankfurt am Main, Germany\\
	\texttt{elu@fias.uni-frankfurt.de}
	\And
	Nenad~Polomac \\
	Institute for Neuroradiology \\
	Frankfurt university hospital\\
	Frankfurt am Main, Germany\\
	\texttt{Nenad.Polomac@kgu.de} \\
	\And
	Iskra~Gacheva\\
	Institute for Neuroradiology \\
	Frankfurt university hospital\\
	Frankfurt am Main, Germany\\
	\texttt{s9819495@stud.uni-frankfurt.de } \\
	\And
	Elke~Hattingen \\
	Institute for Neuroradiology \\
	Frankfurt university hospital\\
	Frankfurt am Main, Germany\\
	\texttt{elke.hattingen@kgu.de} \\
	\And
	Jochen Triesch\\
	Frankfurt Institute for \\
	Advanced Studies (FIAS)\\
	Frankfurt am Main, Germany\\
	\texttt{triesch@fias.uni-frankfurt.de}
}

\begin{document}
	\maketitle
	
	\begin{abstract}
		The application of Deep Learning (DL) for medical diagnosis is often hampered by two problems. First, the amount of training data may be scarce, as it is limited by the number of patients who have acquired the condition to be diagnosed. Second, the training data may be corrupted by various types of noise. Here, we study the problem of brain tumor detection from magnetic resonance spectroscopy (MRS) data, where both types of problems are prominent. To overcome these challenges, we propose a new method for training a deep neural network that distills particularly representative training examples and augments the training data by mixing these samples from one class with those from the same and other classes to create additional training samples. We demonstrate that this technique substantially improves performance, allowing our method to reach human-expert-level accuracy with just a few thousand training examples. Interestingly, the network learns to rely on features of the data that are usually ignored by human experts, suggesting new directions for future research.
	\end{abstract}

	% keywords can be removed
	\keywords{Brain tumor \and Magnetic resonance spectroscopy (MRS) \and Noisy labels \and Deep neural network \and Data augmentation}

	\section{Introduction}
	% epilepsy affects a lot of people -- early identifying EPG or even staging EPG could allow for better effective treatment -- we don't fully understand the EPG, the progression -- there are attempts in discovering EPG biomarkers, we previously tried to identify BL and EPG -- complexity of EPG, it is not like a single biomarker could be responsible -- thanks to recent advances in ML and DL, we are able to learn the features directly in an end-to-end fashion. -- Here, we propose a DNN-based approach. We 
	Modern machine learning (ML) approaches based on deep neural networks have recently obtained impressive results in a range of classification tasks, sometimes even outperforming human experts. These successes are made possible by the combination of 1) better learning algorithms, 2) fast, massively parallel computing hardware including graphics processing units, and 3) the availability of large training data sets. However, in many application domains, such large data sets may simply not exist or be extremely expensive to gather. This problem is particularly severe in certain medical applications, where the numbers of patients may be quite small. Typical data sets may contain only hundreds or thousands of samples, while modern ML approaches often require the estimation of many millions of free parameters. Fitting a model with many free parameters to a small set of training samples will likely lead to over-fitting and poor generalization of the learned model. This problem is aggravated if the training data are corrupted by different kinds of noise, which is often unavoidable in biomedical data. 
	
	\begin{figure}[tb]
		\centering
		\includegraphics[width=0.85\linewidth]{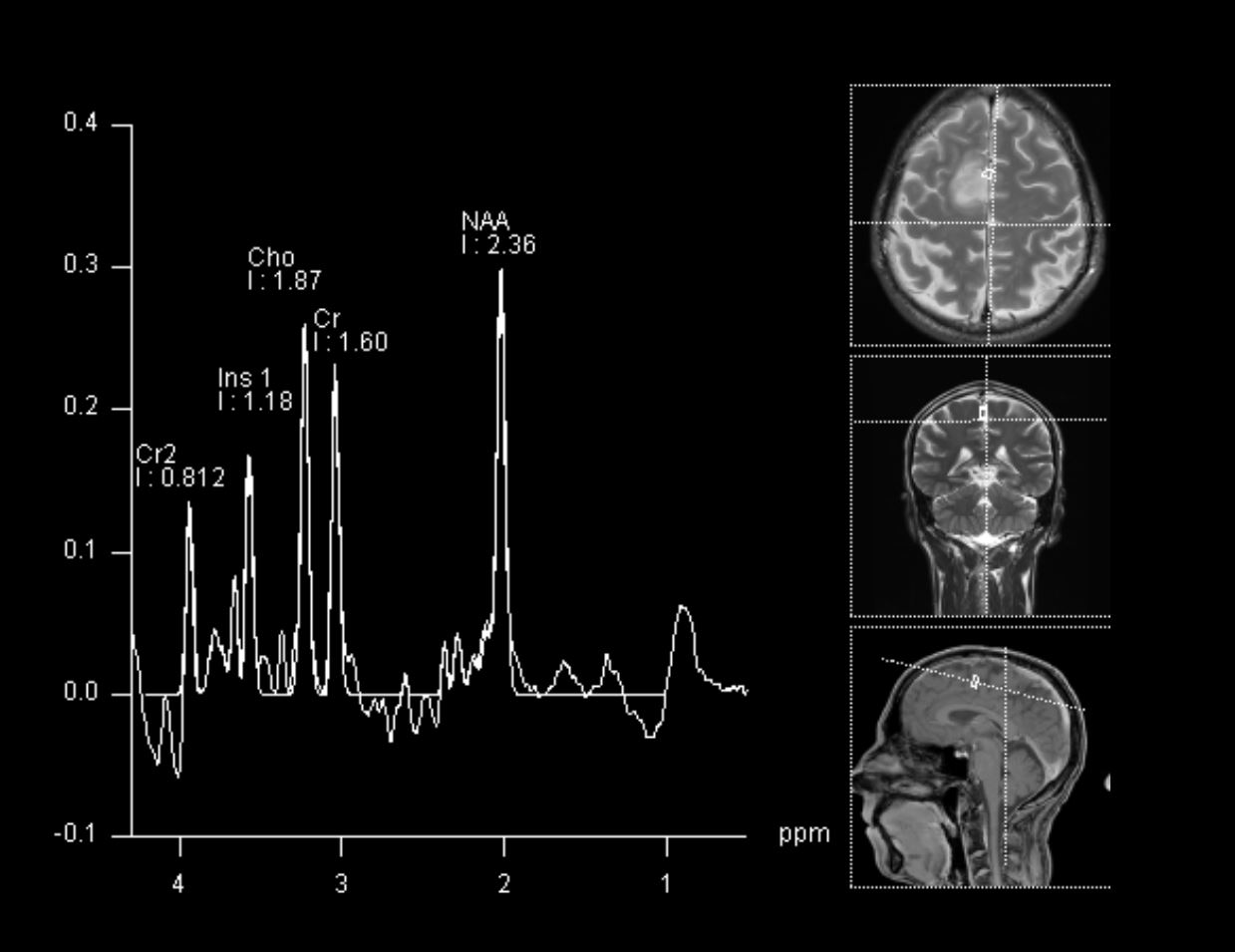}
		\caption{Example MRS spectrum from a healthy patient}
		\label{spectrum}
	\end{figure}
	
	Here, we study the problem of brain tumor detection from magnetic resonance spectroscopy (MRS) data. In clinical practice, MRS is a common tool to identify a brain tumor and distinguish it from other medical conditions. It measures the resonant frequency shift of a chemically bounded hydrogen atom (i.e., a proton), which characterizes different physiological or pathological brain metabolites. There has been increasing interest in MRS for clinical use because of the semiautomatic data acquisition, processing and quantification \cite{kreis2004issues, ranjith2015machine, munteanu2015classification}. An example MRS spectrum from a non-tumor patient is shown in Fig.~\ref{spectrum}. While the interpretation of spectra is traditionally based on the size and location of certain peaks, we here use a novel approach by analysing the pattern of the MR spectrum as a whole in an unbiased fashion with machine learning.

	%{Problems with MRS measures} , noisy, 
	A common problem with in-vivo MRS data is that they are quite noisy. Noise sources range from heterogeneous magnetic susceptibilities of human tissues over baseline distortions of the spectrum \cite{kreis2004issues} to head movement during the procedure. Hence, the quality of spectra may be inadequate to determine precise metabolite concentrations and artefacts may resemble diagnostic features. As an additional problem, during the tissue selection process, due to the indefinable borders of gliomas, spectra from the tumor-affected hemisphere can be falsely labeled as tumor even though they contain healthy brain tissue.
	%The resulting large overlap of the classes can be visualized through k-means clustering of the whole data set using the Euclidean distance metric (Fig.~\ref{crosstabs}). It shows that the spectra from the two classes are largely overlapping. 
	Furthermore, depending on the size of the selected region of interest, the number of samples collected from each patient varies substantially. Such a heterogeneous distribution of the individual training samples impedes the generalization of the learned model --- especially in a leave-one-out (LOO) cross validation scheme. 
	
	\begin{figure*}[tb]
		\centering
		\includegraphics[width=0.95\linewidth]{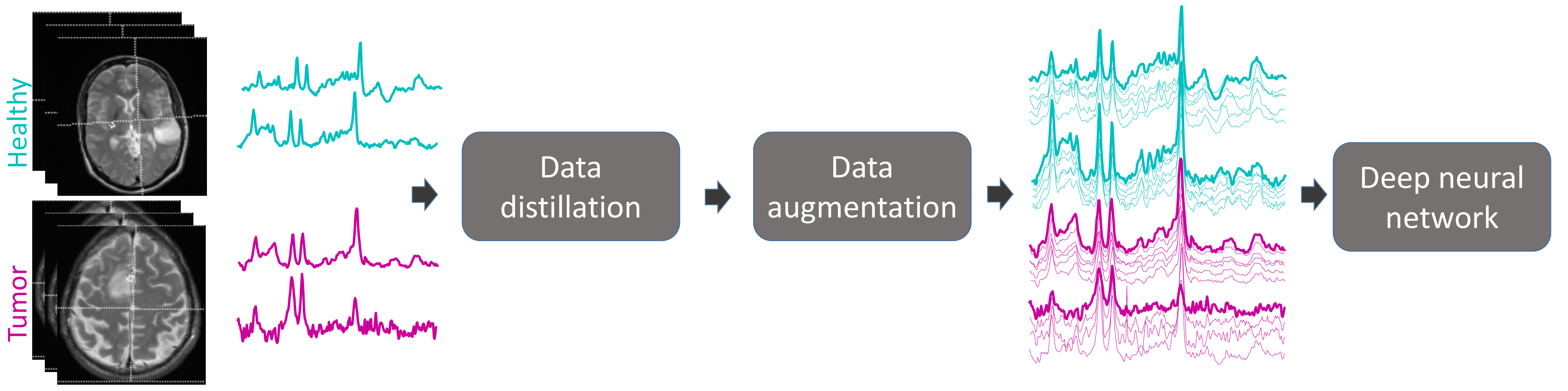}
		\caption{Overview of the proposed approach. MRS spectra from both classes are obtained. A data distillation network automatically selects representative samples that serve as the basis for creating an augmented data set. The augmented data set it used to train a final network for classification.}
		\label{workflow}
	\end{figure*}
	
	Scarcity of training data can be a big hurdle when applying DL methods to medical problems. Data augmentation is a common approach to alleviate this problem. It works by synthesizing new training data from the existing data via a variety of methods reviewed below. Here, we propose a new framework using two separate neural networks: a data distillation network to select representative training examples and a final classification network. In a nutshell, our method works by identifying data points that are ``easy'' to classify through the distillation network. Then, these data samples are used to synthesize a large number of new training data samples for training the final classifier. The new training samples are generated by mixing the easy samples with randomly selected data points from the same or other classes. The proposed framework is illustrated in Fig.~\ref{workflow}. Notably, it does not require human supervision to carefully label a small data set as prototype samples \cite{Li2017a, Lee2018}, but learns directly on the noisy labeled data. We show the benefits of this approach by demonstrating that it outperforms state-of-the-art methods and achieves human-expert-level performance. In sum, we make the following contributions: 
	\begin{itemize}
		\item We propose a framework for tumor classification based on MRS data that combines deep neural networks with a novel data distillation and augmentation procedure to combat scarcity of the training data and labeling noise.
		\item We quantify the performance of human expert neuroradiologists on \textit{tumor}/\textit{healthy} classification from MRS data and demonstrate that our approach achieves human-expert-level performance.
		\item We show that the network uses prominent features in the data that are commonly used in clinical practice, but also considers features that have not yet received much attention by medical professionals, pointing out new directions for future research.
	\end{itemize}

	The remainder of the paper is organized as follows. In Sect.~\ref{s-relate}, we will briefly review state-of-the-art methods for data augmentation and dealing with noisy labels. In Sect.~\ref{s-data}, we present our data set and the data acquisition and preprocessing. In Sect.~\ref{s-method}, we describe the deep neural network architecture and our new data augmentation technique. Sect.~\ref{s-results} presents and discusses our results, showing that our network can achieve human expert-level performance on this task by using the proposed data augmentation approach. Section~\ref{s-conclusion} concludes the paper.
	
	%\IEEEpubidadjcol
	
	\section{Related work}
	\label{s-relate}
	In this section, we provide a brief review of recent research using deep neural networks in medical applications. We focus on the field of oncology and the problems of noisy labels and scarce data.
	
	\subsection{Deep Neural Networks}
	In recent years, DNN-based methods have gained more and more popularity in the healthcare domain and achieved some impressive results \cite{lin2019super, park2019machine,esteva2017dermatologist,haenssle2018man,hannun2019cardiologist,podnar2019diagnosing,capper2018dna,pereira2016brain,munteanu2015classification, ranjith2015machine}. 
	Among different network structures, Convolutional Neural Networks (CNNs) have gained great popularity \cite{simonyan2014very, szegedy2016rethinking, he2016deep}. They are inspired by the information processing mechanism of the visual systems of mammals where individual neurons respond to inputs in a restricted region of the visual field known as their receptive field. In comparison to fully connected neural networks, CNNs have a weight-sharing feature where neurons in different locations have identical receptive fields such that their responses can be calculated via a convolution operation. This design significantly reduces the number of trainable parameters and improves the generalization ability of the network. 
	Ng \textit{et~al.} applied a Deep CNN to electrocardiography (ECG) data in heart disease classification and have achieved better performance than human cardiologists \cite{hannun2019cardiologist}. In \cite{haenssle2018man}, a deep CNN was trained on dermascopic melanoma detection and achieved above-dermatologist performance. In the field of oncology, machine learning methods have obtained promising results on problems such as tumor detection, tumor segmentation, tumor progression, etc. \cite{podnar2019diagnosing, lin2019super, capper2018dna, park2019machine, pereira2016brain, munteanu2015classification, ranjith2015machine}. For examples, Pereira \textit{et~al.}, applied a deep CNN for tumor segmentation from MRI data \cite{pereira2016brain}. Podnar \textit{et~al.} used a machine learning predictive model for the diagnosis of brain tumors from routine blood test results\cite{podnar2019diagnosing}. Machine learning methods applied to MRS data, such as in \cite{ranjith2015machine}, obtained good results in tumor grade classification according to the WHO tumor grade standard. However, learning from a larger cohort with multiple medical conditions only from MRS data has not yet been performed.

	\subsection{Learning from Noisy Labels}
	Noisy labels are ubiquitous in the real world. In this study, noisy labeling refers to observed labels that are incorrect, i.e., due to the labeling procedure, the label assigned to the instance does not represent the class membership. Noisy labels are posing a non-trivial problem in deep model learning when an increasing ability to fit noise is accompanied with deeper layers. Given the ubiquity and importance of coping with noisy labeling, many works have been devoted to combat  this problem \cite{Li2017a, Lee2018, han2018co, smyth2019training, VeitLearning}. One promising direction is to learn from a small set of clean labeled data and then use them to update the network \cite{han2018co, Li2017a, VeitLearning}. Another direction is to design models that could learn directly with noisy labels \cite{han2018co, smyth2019training, RolnickDeep}.
	In \cite{Li2017a}, an auxiliary model is trained with a small but clean data set, which was manually labeled by human experts. Then the knowledge obtained by the auxiliary model is guiding the learning of the primary model in the form of one part of the primary training loss being the imitation loss of the primary model to the auxiliary trained model. 
	Lee \textit{et~al.} proposed a hybrid system, which requires a small set of representative seed instances with precise labels \cite{Lee2018}. Then, the automated noisy label detection is achieved with a deep CNN.
	Veit \textit{et~al.} proposed a semi-supervised learning framework for multilabel image classification that leverages small sets of clean labels in conjunction with large amounts of noisy labels \cite{ VeitLearning}. Small sets of clean labels facilitate the learning of the mapping between noisy and clean labels, which not only reflects the noisy patterns but also the labeling structure.
	Han \textit{et~al.} proposed a co-teaching framework where two DNNs were trained simultaneously \cite{han2018co} with the whole data set. The networks train each other using small-loss training instances, since they are likely to be clean annotations. The intuition is that since two neural networks are different and are equipped with different learning capabilities, during learning they may capture diverse features from the training samples and through the small-loss instances they are filtering out ``clean'' samples for each other. 
	Smyth \textit{et~al.} proposed a DNN-based framework, Training-ValueNet, which evaluates the contribution of one sample to the whole learning process and then discards those that negatively contribute to the learning \cite{smyth2019training}.

	\subsection{Data augmentation}
	%% editing original data, image, blending, 
	A number of techniques have been explored to alleviate the problem of small training data sets. Data augmentation is a very effective way to expand the existing training set with more and diverse data in order to improve the generalization ability and incorporate invariance. Usually, data augmentation methods are domain- and dataset-specific. The fundamental rule of data augmentation is that the meaning of the target samples should be maintained regardless of the augmentation methods applied. The trained model should be reliable enough to predict the same class even when the samples are perturbed. One common class of data augmentation methods especially applicable to image data is based on different data transformations such as cropping, rotating, flipping, shearing, etc. \cite{Krizhevsky2017, simonyan2014very}. Another class of methods is referred to as adversarial training where models are trained with generated adversarial samples \cite{Volpi2018, tran2017bayesian}. In \cite{Volpi2018}, the authors were concerned with the problem of generalizing learning from only one single source distribution to the unseen data domain. They augment the training set with generated adversarial samples. Tran \textit{et~al.} proposed a joint learning scheme where a Bayesian data generator is trained with existing training samples and continuously generates new training samples for further classification \cite{tran2017bayesian}. In \cite{perez2017effectiveness}, images in different styles are generated through a CycleGAN model and then used for further image classification. 
	
	In another line of thinking, one augmentation strategy is blending two or more training samples to generate new ones, though still in its early stage \cite{summers2019improved}. Inoue \textit{et~al.} propose a data augmentation method by mixing randomly selected images from the training set \cite{inoue2018data}. Jaderberg \textit{et~al.} presented a framework for recognizing natural scene text \cite{Jaderberg2014}. In this work, a larger text corpus is generated with font rendering, creating and coloring with a background image-layer, a foreground image-layer, and an optional shadow image-layer. A natural data blending process is applied, where a random crop of an image from the training dataset is blended with each layer of the synthesized image. The three image layers are also blended together randomly to give a single output image. Summers \textit{et~al.} investigated various example-mixing methods in generating new samples and found that all mixing-based data augmentation methods resulted in an improvement of baseline performance \cite{summers2019improved} . In this work, the algorithm learned that mixing several samples of certain classes in a nonlinear way results in an improvement of the generalization ability of the learned model. However, data blending requires more delicate considerations compared to traditional data augmentation methods with various image transformations. Questions such as blending what together, how much of each component should be used, etc., need to be carefully addressed. 
	
	\section{Dataset}
	\label{s-data}
	
	\paragraph{Data acquisition}
	1H-MR-spectroscopy data from 435 patients recorded in the Institute for Neuroradiology of the University Hospital, Frankfurt during the time interval from 01/2009 to 3/2019 were reviewed retrospectively. The spectroscopy was performed on a clinical 3T MR Scanner (Skyra, Siemens Medical Solutions, Erlangen, Germany) using a phased array head coil with 20 arrays with TE = 30 ms; TR~=~1500 ms; flip angle 90$^{\circ}$; scan time of 6:11 min. These patients were suffering from either glial or glioneuronal first diagnosed tumors (the \textit{tumor} group) or other non-neoplastic lesions e.g. demyelination, gliosis, focal cortical dysplasia, enlarged Virchow-Robin spaces or similar (the non-tumor/\textit{healthy} group). The tumor group included all spectra from the tumor-affected hemisphere. The non-tumor group consisted of all spectra from both hemispheres of the patients. As a result, 7442 spectra (3388 non-tumor and 4054 tumor) were selected for further analysis. The obtained MRS examples are saved as column vectors ($288\times1$), shown in Fig.~\ref{spectrum}, where the $y$-axis shows signal intensities of different metabolites, and the $x$-axis represents the chemical shift positions in ppm indicating the various metabolites.

	\section{Methods}
	\label{s-method}
	In this section, we formulate our problem of classifying MRS data collected from patients with and without brain tumors into \textit{tumor} and \textit{healthy} classes with deep neural networks. We first outline the challenges we face in this work and propose solutions. Then, we describe in detail the network structure and the corresponding parameters. To evaluate how well our proposed method works, we construct a performance comparison in a realistic clinical setting with eight neuroradiologists. 
	\subsection{Challenges}
	In our particular problem, we face several challenges regarding the dataset, i.e., noisy labeling, data shortage and imbalanced classes.
	\paragraph{Noisy labeling}
	Infiltrative growth is an important feature of gliomas, which distinguishes them from expansively growing tumors, such as metastases. The real borders of gliomas are indefinable, which can strongly confound the selection and labeling of the voxels from multivoxel spectroscopy. One source of labeling noise is introduced when spectra from the tumor-affected hemisphere are falsely labeled as tumor-containing voxel although they contain healthy brain tissue.
	
	\paragraph{Data Shortage and Class Imbalance}
	A large amount of training data is one of the most essential factors in training DL models successfully. However, as mentioned before, the amount of such MRS data is limited by the number of patients with the medical conditions of interest. Furthermore, as the size of the selected region of interest varies substantially for each patient, the number of samples collected from each patient also varies. Such imbalance can negatively affect the training of a classifier. 
	
	\subsection{Proposed Solutions}
	To tackle the problems mentioned above, we propose the following solutions.
	\paragraph{Data Distillation}\label{disillation}
	To deal with the noisy labeling problem, we propose to distillate the data before they are fed into the deep neural network. In this way, the deep neural network  will start by learning a subset of most representative samples from each class, i.e., ``clean-labeled'' samples \cite{arpit2017closer}. Specifically, we propose a distillation network to collect samples that this network is very certain about during initial training epochs. The intuition is that these samples are highly representative and most likely to be correctly labeled. 
	The detailed data distillation procedure is defined in Algorithm~\ref{distillation}. In the algorithm, $\mathcal{D}=\{{x_1, \dots, x_N}\}$, is the training set with a total number of $N$ training samples. We first train a distillation network, which has the same structure and configuration as the final model. In principle, any type of network can be used for the distillation purpose. The softmax output from the DNN of the $j$-th sample $x_j$ is a two-element vector of classification probabilities over the two possible classes: \textit{healthy} and \textit{tumor}. With a threshold $\theta$, set empirically to $\theta=0.99$, we collect the certain samples where the maximum probability is greater or equal than $\theta$ at the end of each training epoch.
	
	% TODO. future you could collect certain samples with different configuration that capture different aspects of the data.
	% 	\begin{algorithm}[t]
	% 		\KwIn{The training set $\mathcal{D}=\{{x_1, \dots, x_N}\}$}
	% 		\KwOut{$\mathcal{C}$  \quad // collection of certain samples}
	% 		Initialize a DNN\;
	% 		$\mathcal{C} \gets \{\}$  \;
	% 		\For{$E\gets 1$ \KwTo $max_E$}
	% 		{    
	% 			$\mathcal{C}_{E} \gets \{\}$\;
	% 			\For {$j\gets 1$ \KwTo $N$}{   % go through all training samples
	% 				\If{$\max(P(j)) \geq \theta$}{
	% 					$\mathcal{C}_{E}\gets \mathcal{C}_{E} \cup \{x_j\}$\;
	% 				}
	% 			}
	% 			$\mathcal{C} \gets \mathcal{C} \cup \mathcal{C}_{E}$\;
	% 		}
	% 		\caption{Data Distillation}
	% 		\label{distillation}
	% 	\end{algorithm}
	
	{\centering
		\begin{minipage}{.55\linewidth}
			\begin{algorithm}[H]
				\KwIn{The training set $\mathcal{D}=\{{x_1, \dots, x_N}\}$}
				\KwOut{$\mathcal{C}$  \quad // collection of certain samples}
				Initialize a DNN\;
				$\mathcal{C} \gets \{\}$  \;
				\For{$E\gets 1$ \KwTo $max_E$}
				{    
					$\mathcal{C}_{E} \gets \{\}$\;
					\For {$j\gets 1$ \KwTo $N$}{   % go through all training samples
						\If{$\max(P(j)) \geq \theta$}{
							$\mathcal{C}_{E}\gets \mathcal{C}_{E} \cup \{x_j\}$\;
						}
					}
					$\mathcal{C} \gets \mathcal{C} \cup \mathcal{C}_{E}$\;
				}
				\caption{Data Distillation}
				\label{distillation}
			\end{algorithm}
		\end{minipage}
		\par
	}
	
	Here, we add $x_j$ to the certain sample set $\mathcal{C} \subseteq \mathcal{D}$, when the maximum value in $P(j)$ is greater or equal to the threshold $\theta$. As mentioned that the ``clean'' samples are learned first at the initial training process, $max_E$ is the maximum training epoch that is still considered as in the initial training. The set $\mathcal{C}_E$, where $E \in \{{1, 2, \dots, max_E}$\}, is the collection of certain samples from epoch $E$. $\mathcal{C}$ is defined as the union of all $\mathcal{C}_E$: $ \mathcal{C} = \mathcal{C}_1 \cup \ldots \cup \mathcal{C}_E.$ 
	% From the experiment, the later epochs cover the previous samples

	\paragraph{Data Augmentation}
	With an increasing number of distillation epochs, more and more samples are collected. However, with the maximum number of $\textit{E}$, the collected samples are still only a small part of the whole training set. To increase the number of training samples and improve the performance of our primary learning model, we propose a data augmentation method. Let $\mathcal{A}$ denote the augmented set of samples.
	
	A new augmented sample is created according to
	\begin{equation}\label{mix_factor} 
	x^{\mathcal{A}}_i = (1 - \alpha)\cdot x^{\mathcal{C}}_j + \alpha \cdot x^{\mathcal{C}}_k,
	\end{equation}
	where $\alpha \in [0, 1]$ is the mixing weight with a value of 0.5 as default, $x^{\mathcal{A}}_i$ is the newly generated sample, $x^{\mathcal{C}}_j \in \mathcal{C}$ is the target sample which will be augmented, and $x^{\mathcal{C}}_k \in \mathcal{C}$ is the sample that is used to be mixed into the target sample. The total number of augmented samples in set $\mathcal{A}$ divided by that of the set $\mathcal{C}$ is termed the augmentation factor. We propose three different augmentation strategies: augment with the same class (aug-with-same), augment with the opposite class (aug-with-other) and augment with both classes (aug-with-both). Based on the choice of the augmentation strategy, $x^{\mathcal{C}}_k$ could be randomly selected from either class groups or both. During the augmentation, the label of $x^{\mathcal{A}}_i$ is defined as the label of $x^{\mathcal{C}}_j$. To deal with the class imbalance, we apply the method of oversampling the minority class described in \cite{Chawla2002}.
	
	\subsection{Deep Neural Network Structure}
	In our implementation, we apply the residual neural network proposed by He \textit{et~al.} as the backbone \cite{he2016deep}. Residual neural network feature skip-connections, which connect the input of one layer and the pre-activation of another layer skipping multiple layers in between. This structure is usually termed a residual block.  One block usually consists of multiple computational layers such as convolutional or dense layers with batch normalization \cite{ioffe2015batch}, drop-out \cite{srivastava2014dropout}, and a non-linear activation transformation \cite{maas2013rectifier}. The input to the residual block is split into two branches: the main branch with convolution or dense matrix multiplication, batch-normalization, drop-out and the other branch usually with the identity transformation or max-pooling. The combination of the outputs of these two branches is passed through a non-linear activation function as the input of the next block.
	\begin{table}[t]
		\caption{Proposed network structure. The \textbf{Config} column shows the configuration in convolutional and dense layers (filter size $32\times1$ and the number of filters, or the number of units in the dense layer). The number of filters is increased every other block by a factor of 2. Every other block subsamples its input by a factor of 2, indicated by the value of \textbf{Stride}. Here, the batch size at the first dimension is omitted in the output shape column. GAP: global average pooling.}
		\label{structure}
		\centering
		\begin{tabular}{lccc}
			\toprule
			\textbf{Name}     & \textbf{Config}      & \textbf{Stride} & \textbf{Output size}\\
			\midrule
			Conv & $\left[ \begin{array}{cc} 32\times1, & 16  \end{array}\right]$ & 1 &   [batch size, 288, 1, 16] \\
			\midrule
			ResBlock 1 & $\left[ \begin{array}{cc} 32\times1, & 16  \\ 32\times1, & 16  \end{array}\right]$   & 1 &   [batch size, 144, 1, 16]  \\
			\midrule
			ResBlock 2 & $\left[ \begin{array}{cc} 32\times1, & 16  \\ 32\times1, & 16 \end{array}\right]$   & 1 &   [batch size, 144, 1, 16]  \\
			\midrule
			ResBlock 3     & $\left[ \begin{array}{cc} 32\times1, & 32 \\ 32\times1, & 32 \end{array}\right]$ & 2 &  [batch size, 72, 1, 32]     \\
			\midrule
			ResBlock 4     & $\left[ \begin{array}{cc} 32\times1, & 32 \\ 32\times1, & 32 \end{array}\right]$  & 1 &   [batch size, 72, 1, 32]    \\
			\midrule
			ResBlock 5     & $\left[ \begin{array}{cc} 32\times1, & 64 \\ 32\times1, & 64 \end{array}\right]$  & 2 &   [batch size, 36, 1, 64]    \\
			\midrule
			ResBlock 6     & $\left[ \begin{array}{cc} 32\times1, & 64 \\ 32\times1, & 64 \end{array}\right]$  & 1 &   [batch size, 36, 1, 64]    \\
			\midrule
			ResBlock 7     & $\left[ \begin{array}{cc} 32\times1, & 128 \\ 32\times1, & 128 \end{array}\right]$  & 2 &   [batch size, 18, 1, 128]    \\
			\midrule
			ResBlock 8     & $\left[ \begin{array}{cc} 32\times1, & 128 \\ 32\times1, & 128 \end{array}\right]$  &1 &   [batch size, 18, 1, 128]    \\
			\midrule
			GAP          &      &    & [batch size, 128] \\
			\midrule
			Dense        & 2   &     &   [batch size, 2]  \\
			\bottomrule
		\end{tabular}
	\end{table}
	We implement a deep residual neural network with 8 residual blocks following the classic structure from \cite{he2016deep}, including 17 convolutional layers and skip connections. It is inspired by the network architecture in \cite{hannun2019cardiologist}. Each residual block consists of two convolutional layers with batch normalization, drop out and ReLU non-linear activation functions. The convolutional layers have a filter width of $32 \times 1$. Experimenting with different kernel sizes, 32 gives good performance. The number of filters increases by a factor of 2 in every other block starting from 16. There is a sub-sample layer of factor 2 in every other block occurring at the same time when increasing the number of filters. We apply a dropout rate of 0.55 in all blocks. A global average pooling (GAP) layer follows the last convolutional layer to provide further visualization, which is termed a class activation map (CAM) \cite{Zhou2016}. The GAP layer is followed by a soft-max layer, which outputs a probability distribution over the two possible classes. The detailed parameters of the network structure are shown in Table~\ref{structure}.
	
	%The American Medical Association (AMA) has endorsed and adoptd several new policies to support the integration of artificial intelligence in medical practice and training.
	
	\subsection{Visualization through Class Activation Maps}
	Modern DL techniques are often viewed as black-box methods, where the decision making process is difficult to understand for humans. It raises worrying questions and hinders the practical deployment of such techniques. Much effort has been devoted to develop explainable and interpretable DL approaches \cite{Kindermans2018, Yosinski2015a, SebastianOn, Zhou2016}. 
	
	In our work, we apply a GAP layer to reduce the risk of over-fitting and provide further visualization of the network decision making processes. The GAP squashes the output of each feature map with the shape $h\times w\times d$ from the previous layer into one single value with the shape of $1\times 1\times d$ reducing the number of features by $h\times w$ fold. The output of the GAP layer is fed directly to the final classification layer. Intuitively, the GAP operation converts feature maps into weights that represent the ``importance'' of all feature maps, namely the CAMs. An added value of this method is that we can easily trace back the ``importance'' to the input space and visualize how much of each part of the input contributes to the final classification decision.

	\subsection{Quantifying Performance of Human Experts}
	To illustrate how well our proposed method works in comparison to routine clinical diagnostic, a classification task on the same test set is conducted for both the network and human neuroradiologists. Eight experts with different levels of experience in the 1H-MR spectroscopy (from resident to specialist of neuroradiology), were given 844 randomly selected spectra (around 105 per person). They were asked to classify each spectrum as originating from the tumor or from non-tumor tissue reviewing only the spectral lines. They were blinded to any additional information such as T2-weighted images or similar. The overall performance of neuroradiologists is regarded as a collective effort. Inter-rater reliability is not applicable here, since every radiologist received different subsets of the data to classify.

	\section{Results}
	\label{s-results}
	To evaluate performance, we use the receiver operating characteristic (ROC) curve, which is a gold standard to evaluate the discriminative ability of a classifier. It is constructed by varying the classification threshold and calculating the false positive rate (FPR), i.e., 1 - specificity (SPE) with $\text{FPR} = \frac{\text{FP}}{\text{FP} + \text{TN}}$ and the true positive rate (TPR), i.e., sensitivity (SEN) with $\text{TPR} = \frac{\text{TP}}{\text{TP} + \text{FN}}$. The area under the curve (AUC) is a scalar value between zero and one which characterizes the goodness of the classifier. We also compare our results with three baseline methods: a support vector machine (SVM), a random forest (RF), and a plain deep residual CNN (ResCNN) with the same network structure but without distillation and data augmentation. The SVM classifier is implemented with the $sklearn.svm.SVC$ function with default configurations. The RF is implemented with the $sklearn.RandomForestClassifier$ function with 500 trees in the forest and entropy as the measure of the quality of a split. The AUCs of the SVM and the RF are obtained without data distillation and data augmentation as a baseline comparison. 
	
	\begin{table*}[t]
		\caption{Performance measures with default configurations. The performance of the neuroradiologists is computed on one randomly selected cross validation set. The notion ``partial'' in \textbf{Dataset} indicates the test set that was evaluated by the neuroradiologists, whereas ``whole'' refers to averaging across all ten cross validation sets. Results are given as mean $\pm$ standard deviation.}
		\centering
		\begin{tabular}{cccccc}
			\toprule
			&  \textbf{Dataset }& \textbf{SEN}  & \textbf{SPE} & \textbf{AUC} & \textbf{Patient-wise} \\
			&   &   &  &  &  \textbf{accuracy} \\
			\midrule
			Neuroradiologists& partial &  $0.54$  & $\textbf{0.88}$ & $0.68$ & $0.69$\\
			SVM & partial &  $0.66 \pm 0.00$  & $0.58 \pm 0.00$ & $0.62 \pm 0.00$ & $0.57 \pm 0.00$\\
			RF & partial &  $0.63 \pm 0.01$  & $0.60 \pm 0.01$ & $0.61 \pm 0.01$ & $0.63 \pm 0.01$\\
			Plain ResCNN & partial &  $\textbf{0.74} \pm\textbf{ 0.03}$  & $0.58 \pm 0.04$ & $0.68 \pm 0.02$ & $0.69 \pm 0.15$\\	
			Proposed model &partial &  $0.73 \pm 0.20$  & $0.76 \pm 0.30$ & $\textbf{0.77} \pm \textbf{0.08}$ & $\textbf{0.76} \pm \textbf{0.15}$\\		
			\midrule				
			SVM &whole &  $0.69 \pm 0.08$  & $0.61 \pm 0.17$ & $0.63 \pm 0.05$ & $0.67 \pm 0.07$\\
			RF &whole &  $0.70 \pm 0.08$  & $0.60 \pm 0.15$ & $0.64 \pm 0.05$ & $0.70 \pm 0.09$\\
			Plain ResCNN & whole &  $0.70 \pm 0.03$  & $0.60 \pm 0.03$ & $0.69 \pm 0.04$ & $0.67 \pm 0.50$\\
			Proposed model & whole &  $\textbf{0.71} \pm \textbf{0.05}$  & $\textbf{0.69} \pm \textbf{0.07}$ & $\textbf{0.73} \pm \textbf{0.07}$ & $\textbf{0.73} \pm\textbf{0.30}$\\
			\bottomrule
		\end{tabular}
		\label{without}
	\end{table*}
	
	\subsection{Training procedure}
	The ability of the classifier to generalize to new previously unseen patients is of great clinical importance. Therefore, we apply a 10-fold leave-subjects-out cross validation scheme. To be specific, we divide the patient list into 10 sub-lists each with around 40 patients. In each cross validation set, we withhold the data from the patients of one sub-list, while we train and validate on the data from the other sub-lists. The patient-wise accuracy is computed in each leave-out test set. For each patient, the classification probability of all voxels are averaged to get the probabilities of each class. Then, the patient-wise diagnosis is obtained as the class that has the highest probability. The patient-wise accuracy is defined by the number of correct patient-wise diagnoses divided by the total number of patients in that set. 
	We randomly select one cross validation set which consists of 844 spectra from 40 patients for the final test against human neuroradiologists. 
	The network is trained with randomly initialized weights using the Adam optimizer with default parameters $\beta_1=0.9$ and $\beta_2=0.999$ and a mini-batch size of 32. If not specified, the default parameters for data distillation and data augmentation are as follows: the primary learning model is trained with the certain samples collected from the first epoch up to the fifth epoch ($E=5$) from the data distillation network; the training data is augmented with both classes five-fold with the mixing weight $\alpha=$0.5. With our network configuration, at the epoch $E=5$, the average number of collected certain samples is around 2000, providing a total of around 10,000 samples with five-fold augmentation.
	
	To get an average performance of the effect of the proposed distillation process, we train the whole framework twice, i.e., a distillation network, which collects certain samples and a primary classifier with proposed data augmentation on all 10 cross-validation sets, with different random seeds. The results are averaged across the 10 cross-validation sets as well as the two runs. The overall performance is reported in Table.~\ref{without}. It shows that our proposed method slightly outperforms the human neuroradiologists.
	
	\begin{figure*}[tb]
		\centering
		\includegraphics[width=0.95\linewidth]{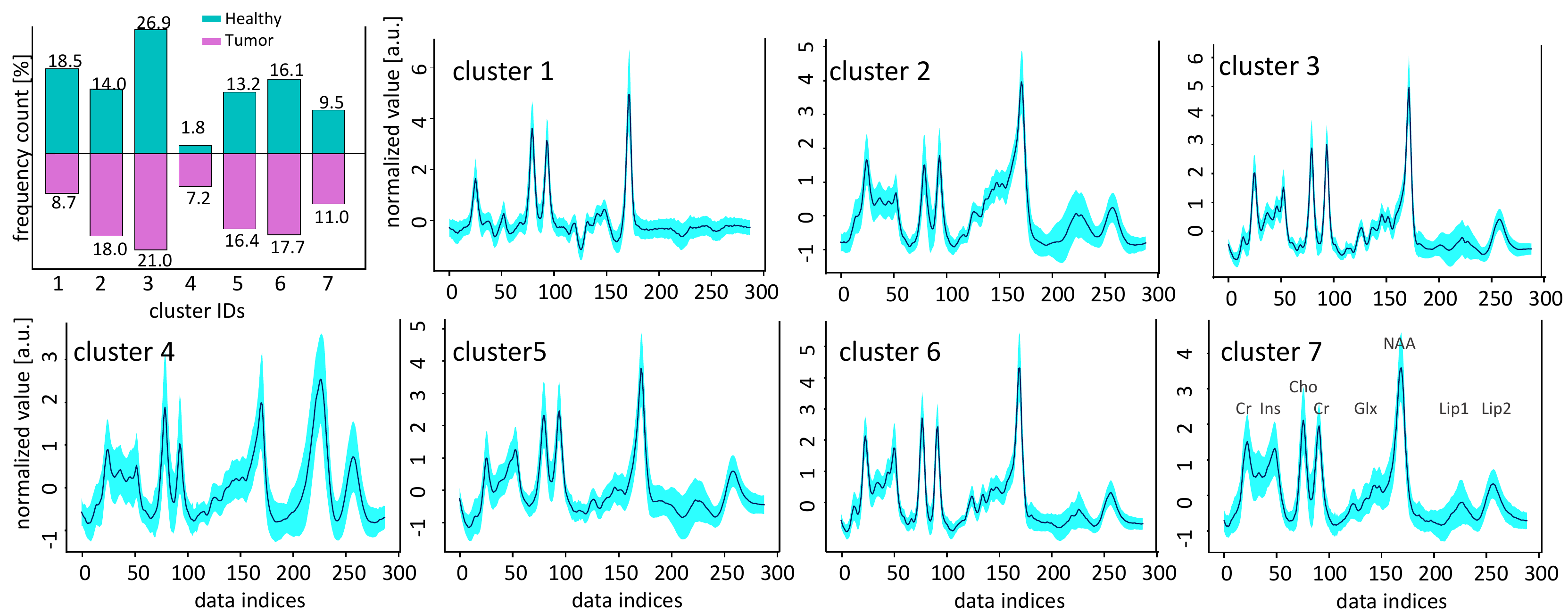}
		\caption{K-means clustering on the whole data set. \textbf{A}. Cross-tab relation of the clustering results. \textbf{B-H}. Mean spectra of each cluster (dark blue) with standard deviation (cyan). }
		\label{clustering}
	\end{figure*}

	\begin{figure*}[t]
		\centering
		\includegraphics[width=0.98\linewidth]{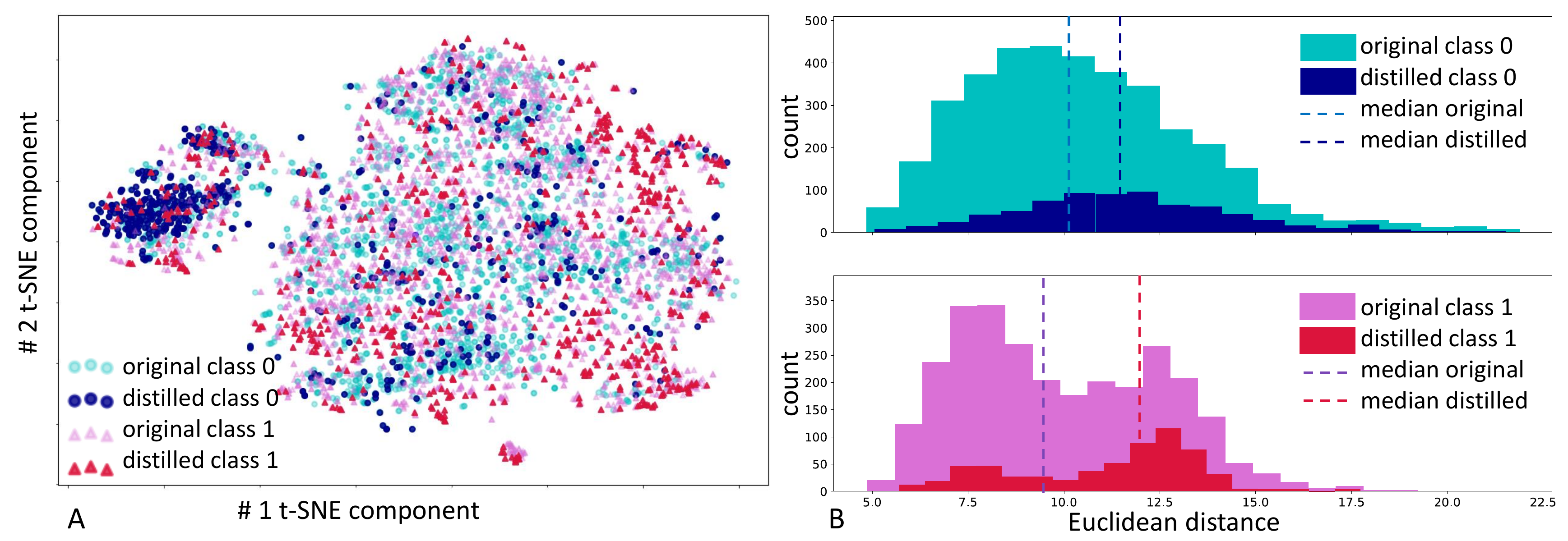}
		\caption{Distillation effect visualization. A: 2-D t-SNE visualization of all training samples  (light blue and light pink) and distilled samples from both classes (blue and red). B: Euclidean distance histogram of all training samples and distilled samples from both classes. Dashed lines are the medians in each class with and without distillation. }
		\label{hist}
	\end{figure*}
	
	\subsection{K-Means Clustering of the Data}
	To get an overview of the data we use in this task, we performed k-means clustering on the whole data set $\mathcal{D}$, which has 7442 spectra (3388 healthy and 4054 tumor). The euclidean distance is used as the criterion to cluster the data. The number of clusters is determined by the elbow point in the inertia curve \cite{kodinariya2013review}, where the within cluster distance does not decrease significantly with an increasing number of clusters (a number of seven is chosen in this study). We did find a large overlap between two classes as we expected.
	
	The clustering results are shown in Fig.~\ref{clustering}. The cross-tab relation, which is a frequency count of one variable (\textit{healthy} or \textit{tumor}) in each cluster is shown in Fig.~\ref{clustering}. A. For example, cluster 1 contains 18.5\% of the healthy spectra and 8.7\% of the tumor spectra We can see that 1) there are samples from healthy and tumor group in every cluster, 2) there are roughly equal amounts of healthy and tumor samples in clusters 2, 3, 5, 6 and 7, 3) the majority of samples in cluster 1 are showing typical features of \textit{healthy} (no Lip1 or Lip2 concentration \cite{Fan2006Magnetic}) and those of cluster 4 are mainly typical \textit{tumor} (high Lip peaks, an elevated Cho peak, high Glx region, etc. \cite{RaeRE}), and 4) the majority of the spectra are neither typical healthy nor tumor, rather somewhere in between. The positions of typical metabolites are demonstrated in Fig.~\ref{CAM}-A. The mean spectra of those clusters illustrate commonly applied clinical assessment criteria:  in healthy tissues, there is a dominant peak at NAA and almost no mobile lipids to be detected since they are mostly confined to the membrane \cite{RaeRE}. In tumor tissues, there are elevated Cho and Lip peaks \cite{Fan2006Magnetic}. A median to high Cho peak with easily visible Cr peaks can contribute to the identification of a tumor \cite{Fan2006Magnetic}. The clustering results support our argument that the labeling process is noisy, so the spectra from both classes are largely mixed with each other.

	\subsection{Distillation Analysis}
	To better understand the effect of the distillation process, we investigate what samples are collected through the first distillation network. To simplify the visualization, we randomly select one training set and its corresponding distilled set after five epochs of training. First, we present a visualization of the whole training set through a 2-D projection via t-SNE \cite{maaten2008visualizing}, shown in Fig.~\ref{hist}A. Light blue and light pink dots represent original training samples from class 0 and class 1, respectively. Then, we highlight the samples collected by the distillation process with darker colors (blue and red for class 0 and class 1, respectively). One can observe that the overall distributions of the whole training samples from both classes largely overlap with each other. However, the distillation process drives the distributions of both classes further apart. For example, the collected \textit{healthy} data samples lie mostly on the left side of the 2-D projection and the collected \textit{tumor} samples reside away from the center of the projetion of the \textit{healthy} class. Second, we quantify the distribution of Euclidean distances of samples from both classes with respect to the center of the opposite class in the original data space. The histograms of the Euclidean distances are shown in Fig.~\ref{hist}B. Indeed, the distillation process collects samples that are further away from the other class's center. 
	
	\subsection{Data Augmentation}
	In this section, we discuss different effects on learning resulting from different options including the mixing weight $\alpha$, the augmentation factor, the index of the last source epoch from which we collect the certain samples and the three augmentation strategies (aug-with-same, aug-with-other and aug-with-both). 
	%As mentioned in section \ref{s-method}, we collected the certain samples during the training and validation processes of the distillation network at each training epoch. With these collected samples, we then train a primary learning network with data augmentation.
	We measure the AUC of the ROC curve with different parameter options for different augmentation strategies. The results are averaged across all 10 cross-validation sets with two different initial distillation networks. 
	
	Adding noise to augment data is a common practice in image data enrichment. Here, we also report results for the case when Gaussian noise is added to augment the data (noise augmentation). We explore different hyper-parameters such as noise amplitude and augmentation factor, and report the performance under the parameters that yielded the best result during the exploration. 
	
	% 	\begin{figure}[tb]
	% 		\centering
	% 		\includegraphics[width=0.95\linewidth]{auc_as_mix.pdf}
	% 		\caption{The AUC as a function of the mixing weight $\alpha$ (from 0.05 to 0.95) for three different augmentation methods. The default augmenting factor is 5 and the index of the last source epoch is $E=5$. Error bars represent one standard deviation across all cross-validation sets. RF: random forest, SVM: support vector machine.}
	% 		\label{auc_mix}
	% 	\end{figure}
	
	\begin{figure}[!tbp]
		\centering
		\begin{minipage}[b]{0.48\textwidth}
			\includegraphics[width=\textwidth]{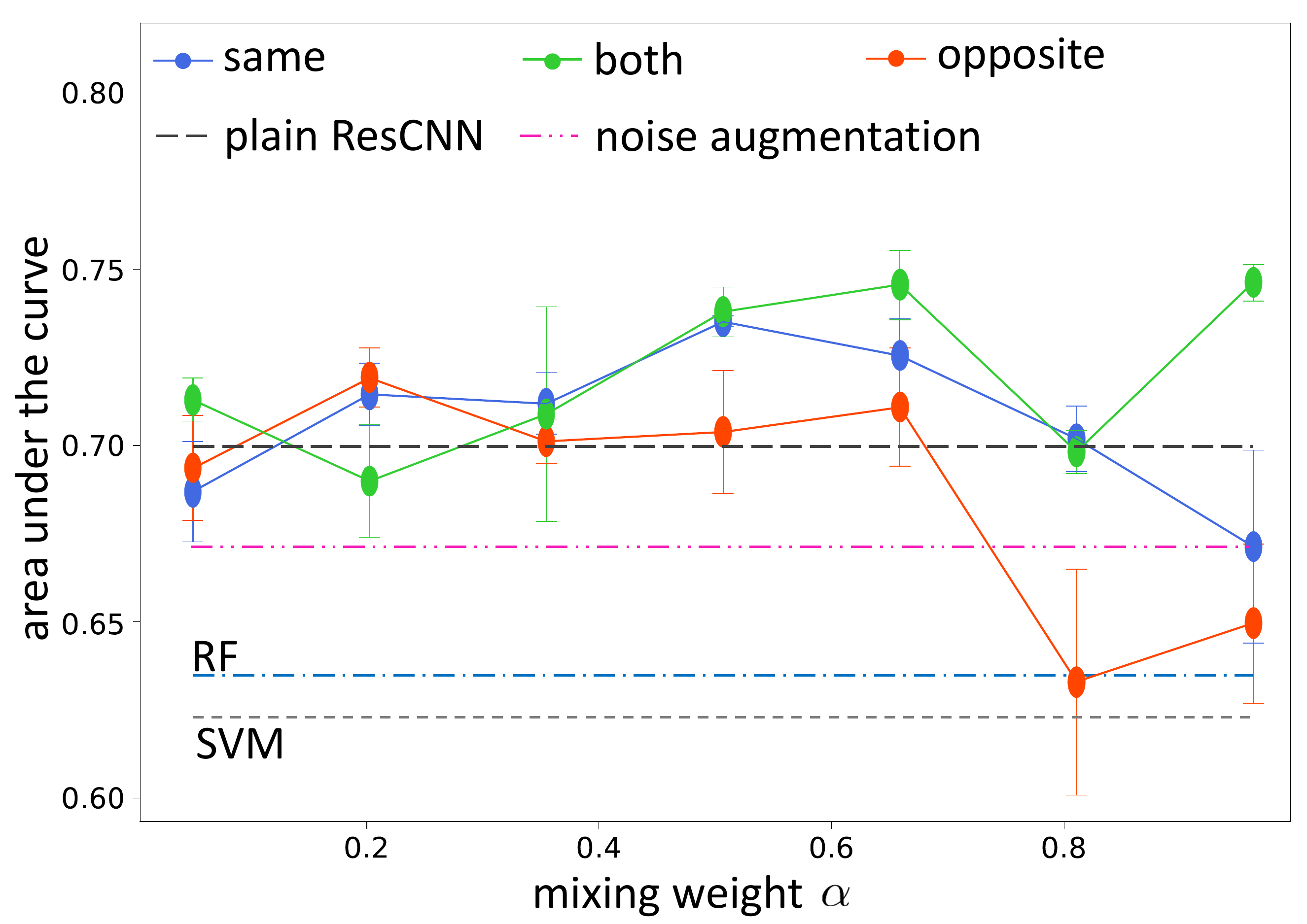}
			\caption{The AUC as a function of the mixing weight $\alpha$ (from 0.05 to 0.95) for three different augmentation methods. The default augmenting factor is 5 and the index of the last source epoch is $E=5$. Error bars represent one standard deviation across all cross-validation sets. RF: random forest, SVM: support vector machine.}
			\label{auc_mix}
		\end{minipage}
		\hfill
		\begin{minipage}[b]{0.48\textwidth}
			\includegraphics[width=\textwidth]{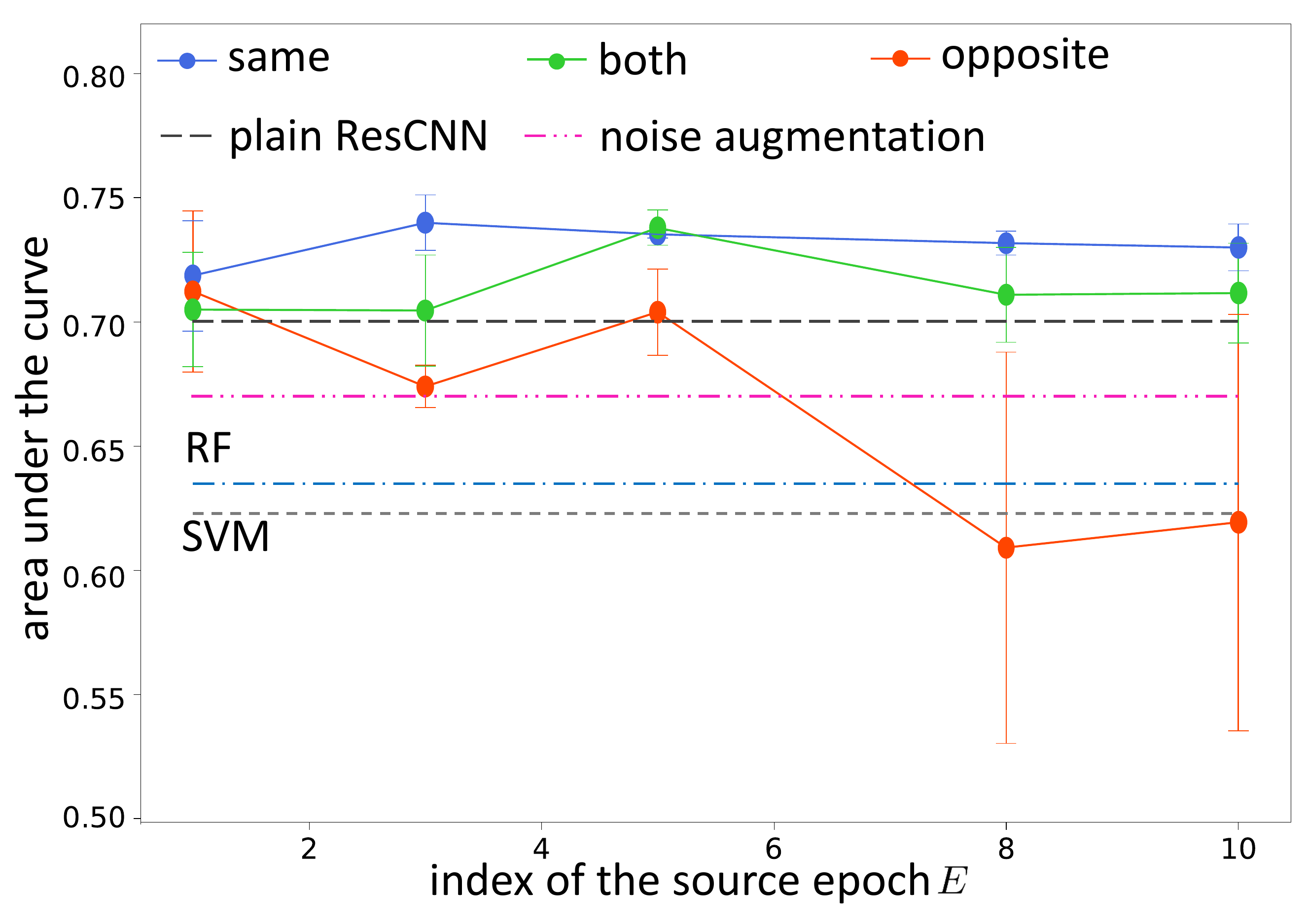}
			\caption{The AUC as a function of the source epoch index $E$ whose certain samples we used for data distillation for the three different augmentation strategies. The default augmenting factor is 5 and the mixing weight is $\alpha = 0.5$. Error bars represent one standard deviation across all cross-validation sets.}
			\label{auc_epoch}
		\end{minipage}
	\end{figure}
	
	In Fig.~\ref{auc_mix},	we show the AUC as a function of the mixing weight $\alpha$ for different augmentation scenarios. We note a number of observations. First, the plain ResCNN without any data augmentation already outperforms the simple SVM and RF classifiers (horizontal lines). Second, the common data augmentation technique of adding noise to the data points actually deteriorates performance compared to the plain ResCNN. Third, of the three proposed data distillation and augmentation approaches, the ``other'' method performs comparably to the plain ResCNN, sometimes giving slightly better or slightly worse results depending on the mixing weight. Fourth, best performance is achieved with the ``both'' and ``same'' methods, which show improved performance compared to the plain ResCNN over a range of intermediate mixing weights.
	%In this experiment, we vary the factor $\alpha$ following Eq.~(\ref{mix_factor}) while keeping the original label. It is also true even in the aug-with-oppo case and $\alpha$ is larger than 0.5, which means there is more of $S_{aug}$ than $S_{target}$ in $S_{new}$ but with the label of $S_{target}$. The AUC shows a slight increase with an increasing $\alpha$ up to 0.65 in aug-with-same and aug-with-both scenarios. In the aug-with-oppo case, it exhibits inferior performance with the highest standard deviation compare to other two scenarios. Interestingly, we do not see a drop rather a slight increase in the AUC when $\alpha$ is 0.95 in the aug-with-both case since there are half of the samples are augmented with the opposite class while keep their original labels. It could due to the fact that the data are noisy since there are ``tumor-looking'' spectra in healthy group and vice versa. Augmenting samples with randomly selected ones from both classes could imitate the statistics in the test data set.
	
	% 	\begin{figure}[tb]
	% 		\centering
	% 		\includegraphics[width=0.95\linewidth]{auc_as_epoch.pdf}
	% 		\caption{The AUC as a function of the source epoch index $E$ whose certain samples we used for data distillation for the three different augmentation strategies. The default augmenting factor is 5 and the mixing weight is $\alpha = 0.5$. Error bars represent one standard deviation across all cross-validation sets.}
	% 		\label{auc_epoch}
	% 	\end{figure}
	
	In Fig.~\ref{auc_epoch}, we show the AUC as a function of the index of the source epoch $E$, when the collection of distilled samples stops (see Algorithm ~\ref{distillation}). Here, $E = 1, \dots, max_E$, where $max_E$ is set to be 10. The mixing weight $\alpha$ was 0.5 and the augmentation factor was 5. In this task, 
	The AUC shows no clear preference among augmenting with ``same''and ``both'' cases, which are always superior to the plain ResCNN method. However, the average performance in the ``other'' case drops (with large standard deviation) as $E$ increases.

	% 	\begin{figure}[tb]
	% 		\centering
	% 		\includegraphics[width=0.95\linewidth]{auc_as_factor.pdf}
	% 		\caption{The AUC as a function of the augmentation factor in three different augmentation strategies during the test. The default source epoch index is 5 and the mixing weight is 0.5.  Error bars represent one standard deviation across all cross-validation sets.}
	% 		\label{auc_fold}
	% 	\end{figure}
	
	\begin{figure}[!tbp]
		\centering
		\begin{minipage}[t]{0.48\textwidth}
			\includegraphics[width=\textwidth]{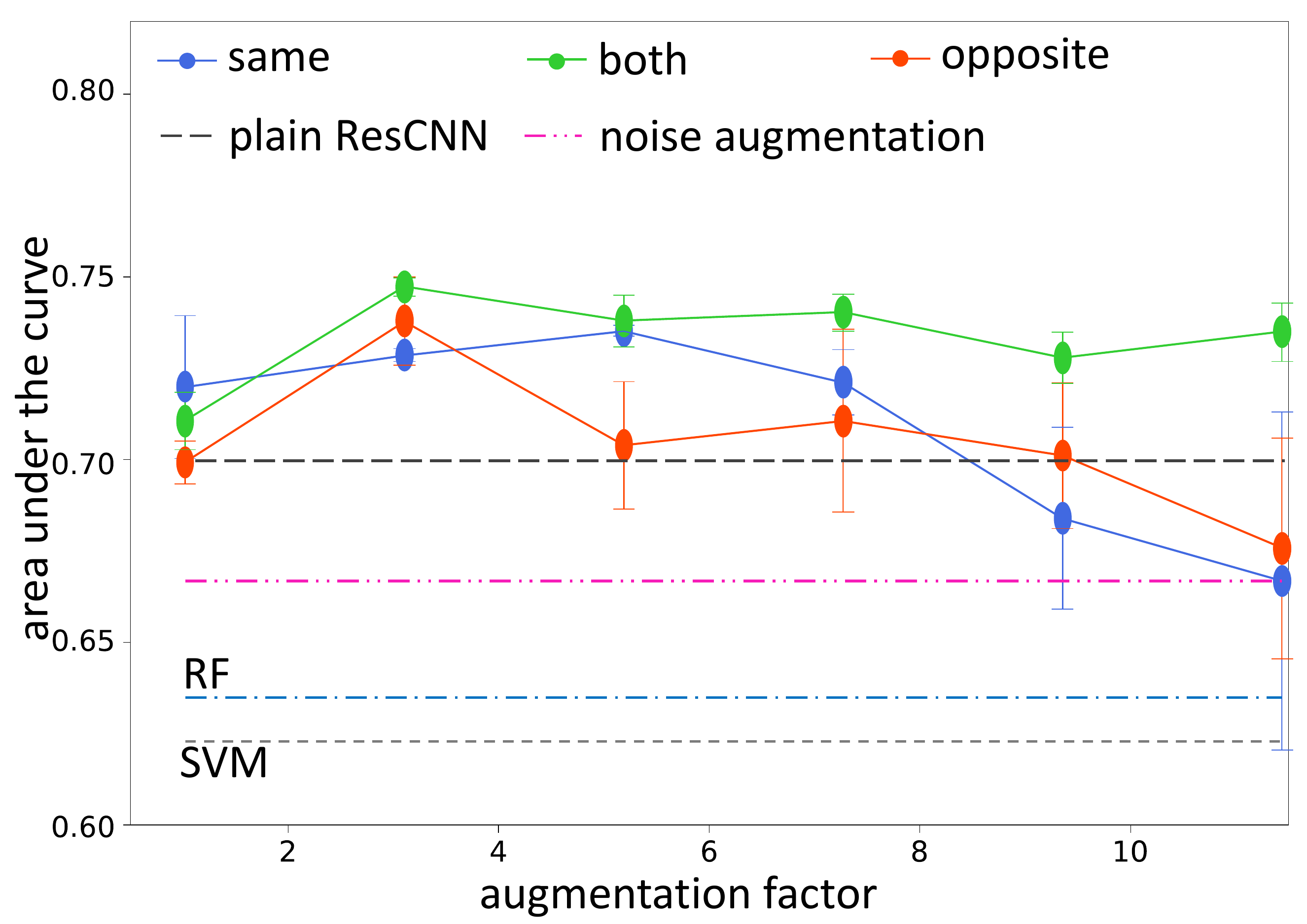}
			\caption{The AUC as a function of the augmentation factor in three different augmentation strategies during the test. The default source epoch index is 5 and the mixing weight is 0.5.  Error bars represent one standard deviation across all cross-validation sets.}
			\label{auc_fold}
		\end{minipage}
		\hfill
		\begin{minipage}[t]{0.48\textwidth}
			\includegraphics[width=\textwidth]{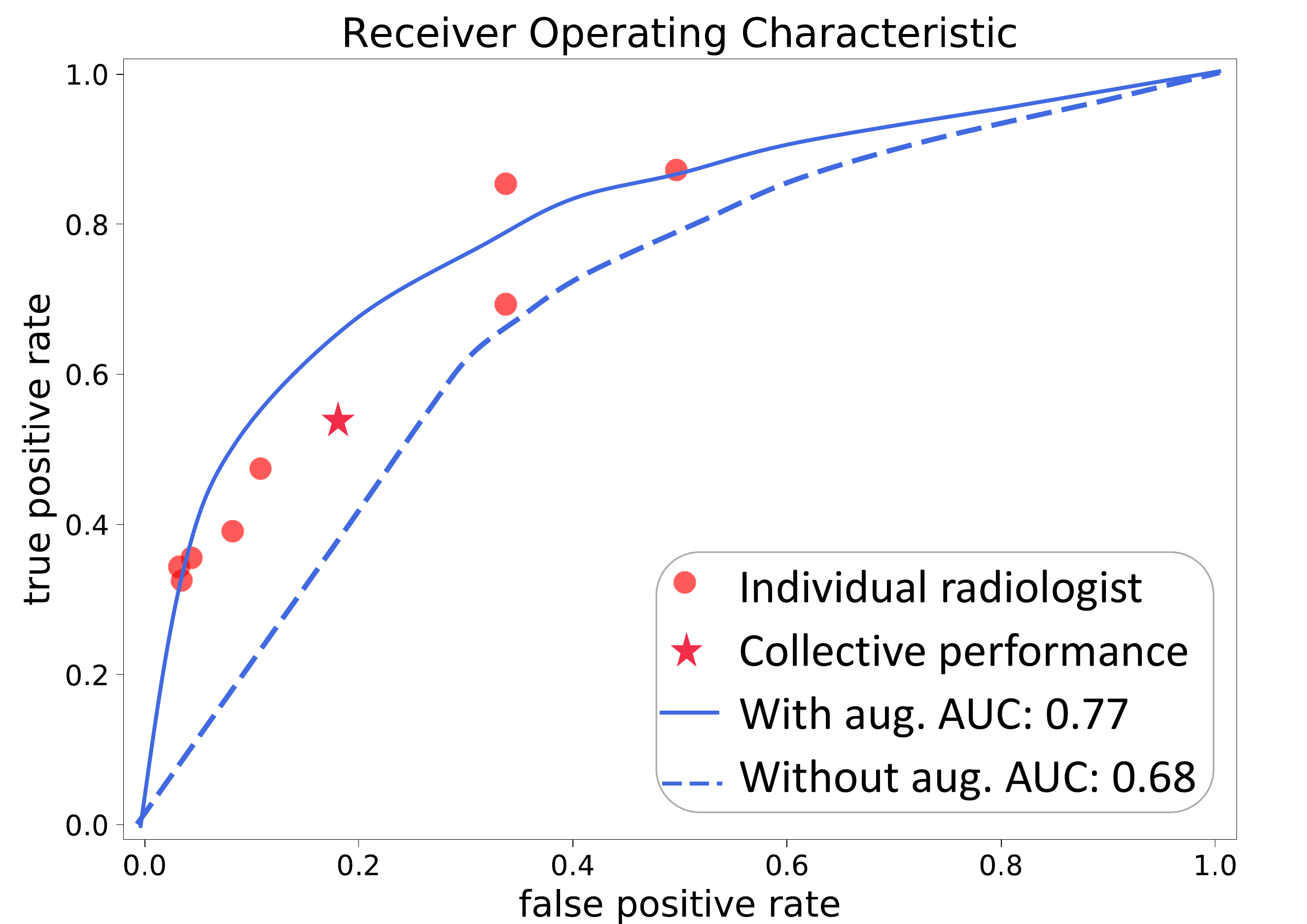}
			\caption{Comparison of the model and neuroradiologists on one randomly selected cross validation set. The individual and collective performance of neuroradiologists are shown as red dots and red star, respectively. The average ROC curve of the plain ResCNN model and our proposed model with default augmentation parameters (aug-with-both method, augmentation factor is five, and the mixing weight is $ \alpha = 0.5$) are depicted in dashed and solid blue lines, respectively.}
			\label{comparison}
		\end{minipage}
	\end{figure}
	
	The AUC as a function of the augmentation factor is shown in Fig.~\ref{auc_fold}. The index of the source epoch $E$ is five, and the mixing weight is $\alpha = 0.5$. It shows that with an increasing augmenting factor up to 3 the result shows a slight increase of the performance. Then, when it goes beyond five it shows a slight decrease in aug-with-\textit{same} and aug-with-\textit{other} augmentation scenarios. 
	%It is possible that when augmenting too many factor, it introduces more variance than invariance, which might hinder the network's learning. 
	
	\subsection{Human vs. Machine}
	To assess how well our proposed method works in a more realistic clinical setting, we compared it to human neuroradiologists on one randomly selected test set. The result is shown in Fig.~\ref{comparison}. 	
	The performance of each individual neuroradiologist is denoted as a red dot, the collective performance is shown as a red star. The model without data augmentation has an AUC of 0.71 (dashed blue line). Our method achieves an AUC of 0.77 (solid blue), which encompasses most of the neuroradiologists in the ROC plot. It shows that our proposed method slightly outperforms the group of neuroradiologists as a whole (AUC: 0.77 vs. 0.68; sensitivity 0.73 vs. 0.54; accuracy: 0.76 vs. 0.69). 
	
	% 	\begin{figure}[tb]
	% 		\centering
	% 		\includegraphics[width=0.95\linewidth]{comparison.pdf}
	% 		\caption{Comparison of the model and neuroradiologists on one randomly selected cross validation set. The individual and collective performance of neuroradiologists are shown as red dots and red star, respectively. The average ROC curve of the plain ResCNN model and our proposed model with default augmentation parameters (aug-with-both method, augmentation factor is five, and the mixing weight is $ \alpha = 0.5$) are depicted in dashed and solid blue lines, respectively.)}
	% 		\label{comparison}
	% 	\end{figure}
	
	\subsection{Feature Visualization}
	As described in section \ref{s-method}, we apply a GAP layer after the convolutional layers to prevent over-fitting and benefit from the possibility of visualizing class activation maps. These show how the network is making the final decision by assigning different weights, which can be interpreted as ``importance'', to different regions in the input data. 
	
	\begin{figure}[th]
		\centering
		\includegraphics[width=0.85\linewidth]{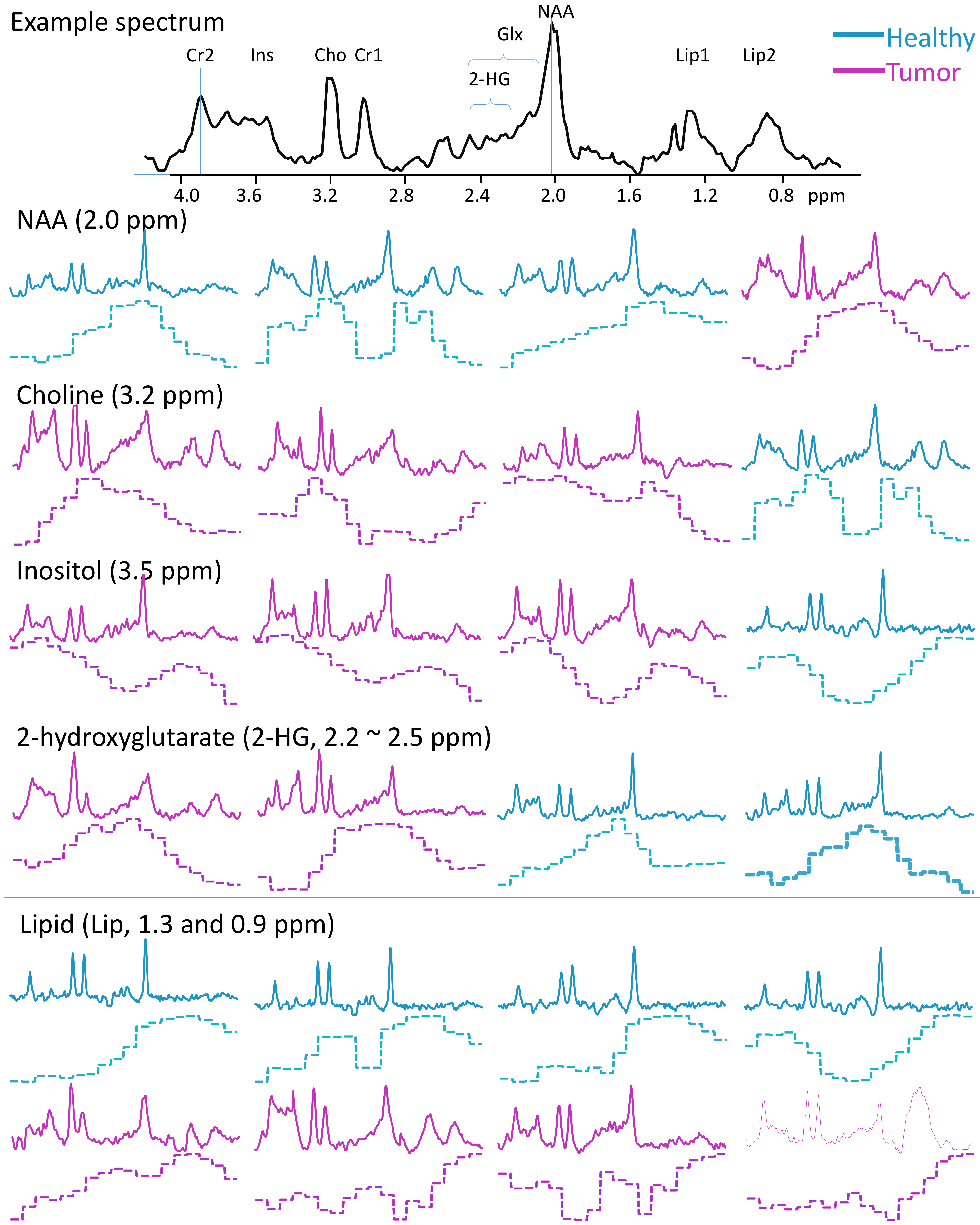}
		\caption{Class activation maps of examples from both classes during test. At the top is an example where various metabolite peaks have been marked. Examples from class \textit{healthy} and \textit{tumor} are color coded in green and purple, respectively. Solid lines are original examples and the dashed line below is the corresponding CAM. Cr1, Cr2: creatine, Ins: myo-inositol. Cho: choline. NAA: n-acetylaspartic acid. Glx: glutamine, Lip: lipid}
		\label{CAM}
	\end{figure} 
	
	In Fig.~\ref{CAM}, we show some examples of CAMs with original MRS samples from both classes. The results show that the CAMs vary with regard to specific samples. To interpret these CAMs, one must not only focus on the highest peak but rather the overall shape together with the original spectra. Note that the ``importance" does not reflect whether the signal intensity of the corresponding metabolite is high or low. The co-occurrence of high ``importance" regions provides insights in the CAM interpretation. We can see that the network considers various common metabolites during the classification. Interestingly, for the \textit{healthy} class the network also pays more attention to the plateau left of the dominant NAA peak. In the \textit{tumor} spectra, this part of the spectrum appears as a rising slope, and represents the oncometabolite 2-hydroxyglutatat \cite{Andronesi,Yen} as well as tumor associated metabolites like glutamine \cite{Li}. The Ins peak together with Cr2 and Lip regions are highly interesting. A high Ins peak with above-baseline Lip peaks highly suggests tumor presence and a low Ins concentration with almost no free lipids suggests the \textit{healthy} class \cite{Lipid}. In cases where a high ``importance" is assigned to the Cho region, the \textit{tumor} spectra show a high Cho peak flanked by other tumor-associated metabolite peaks (glycine, myo-inositol) \cite{naaVScholin,Hattingen2009}. On the other hand, the \textit{healthy} group shows a similar or smaller Cho peak as the Cr1 peak.

	\section{Conclusion}
	\label{s-conclusion}
	In this paper, we present a DNN-based framework, which achieves above human-level performance on a realistic clinical task of classifying tumor and non-tumor tissues based on MRS data. We construct an effective data cleaning and augmentation framework consisting of two steps: 1) a data distillation network to clean noisy labeled data, 2) a data augmentation process, which enlarges the data set acquired in the first step for training a primary neural network for the final classification. Due to its generality, this data augmentation method could be used in various other research domains. By exploring various configurations of the proposed data augmentation method, we further demonstrate that data augementation by mixing samples from both classes is more stable and yields better results. A deep residual neural network is used as the primary learning model and a global average pooling (GAP) layer at the end of all convolutional layers provides us with a visualization of how much each part of the input contributes to the final classification decision. Our proposed framework outperforms neuroradiologists on sensitivity and patient-wise diagnosis accuracy with an area under the ROC curves of 0.77. With an improved capability of coping with noisy labeling and the scarcity of the training data, we believe that the framework proposed in this work could improve clinical practice, ultimately leading to more effective and accurate diagnosis of brain tumors in patients.
	
	\section*{Acknowledgment}
	This work is supported by the China Scholarship Council (No. [2016]3100), the LOEWE Center for Personalized Translational Epilepsy Research (CePTER), and the Johanna Quandt Foundation. Special thanks to Charles Wilmot for inspiring discussions. Furthermore a particular appreciation goes to Marija Radović for her ideas on automating the data export.

	\bibliographystyle{unsrt}
	\bibliography{ref-meta}


	% \section{Class Scores}
	% Class scores of other rats
	% \begin{figure*}[t]
	% 	\centering
	% 	\includegraphics[width=0.9\linewidth]{class-score-other-rats.pdf}
	% 	\caption{Class scores}
	% 	\label{fig:class-score-other}
	% \end{figure*}
	% \subsection{Class Typical Signals}
	
\end{document}